\documentclass{article}

\usepackage{microtype}
\usepackage{graphicx}
\usepackage{subfigure}
\usepackage{booktabs} 

\usepackage{hyperref}

\usepackage[accepted]{icml2025}

\usepackage{amsmath}
\usepackage{amssymb}
\usepackage{mathtools}
\usepackage{amsthm}

\usepackage[capitalize,noabbrev]{cleveref}

\theoremstyle{plain}

\theoremstyle{definition}

\theoremstyle{remark}

\usepackage[textsize=tiny]{todonotes}

\usepackage{multirow} 
\newcommand\MF{L2H-CC}
\newcommand\MS{H2L-CC}
\begin{document}

\twocolumn[
\icmltitle{High Dynamic Range Novel View Synthesis with Single Exposure}




\begin{icmlauthorlist}
\icmlauthor{Kaixuan Zhang}{njust}
\icmlauthor{Hu Wang}{uestc}
\icmlauthor{Minxian Li}{njust,skl}
\icmlauthor{Mingwu Ren}{njust,skl}
\icmlauthor{Mao Ye}{uestc}
\icmlauthor{Xiatian Zhu}{uos}
\end{icmlauthorlist}

\icmlaffiliation{njust}{Nanjing University of Science and Technology.}
\icmlaffiliation{uestc}{University of Electronic Science and Technology of China.}
\icmlaffiliation{uos}{University of Surrey.}
\icmlaffiliation{skl}{State Key Laboratory of Intelligent Manufacturing of Advanced Construction Machinery.}

\icmlcorrespondingauthor{Minxian Li}{minxianli@njust.edu.cn}

\icmlkeywords{Machine Learning, ICML}

\vskip 0.3in
]



\printAffiliationsAndNotice{} 
\begin{abstract}
High Dynamic Range Novel View Synthesis (HDR-NVS)
aims to establish a 3D scene HDR model from Low Dynamic Range (LDR) imagery. Typically, multiple-exposure LDR images are employed to capture a wider range of brightness levels in a scene, as a single LDR image cannot represent both the brightest and darkest regions simultaneously. While effective, this multiple-exposure HDR-NVS approach has significant limitations, including susceptibility to motion artifacts (\textit{e.g.}, ghosting and blurring), high capture and storage costs.
To overcome these challenges, we introduce, for the first time, the  {\em single-exposure} HDR-NVS problem, where only single exposure LDR images are available during training. 
We further introduce a novel approach, Mono-HDR-3D, featuring two dedicated modules formulated by the LDR image formation principles, one for converting LDR colors to HDR counterparts and the other for transforming HDR images to LDR format so that unsupervised learning is enabled in a closed loop.
Designed as a meta-algorithm, our approach can be seamlessly integrated with existing NVS models. Extensive experiments show that Mono-HDR-3D significantly outperforms previous methods.
Source code is released at \url{https://github.com/prinasi/Mono-HDR-3D}.
\end{abstract}

\section{Introduction}
\label{introduction}

\begin{figure}[htp]
\centering
\setlength{\abovecaptionskip}{-0.5cm}
\includegraphics[width=8.2cm]{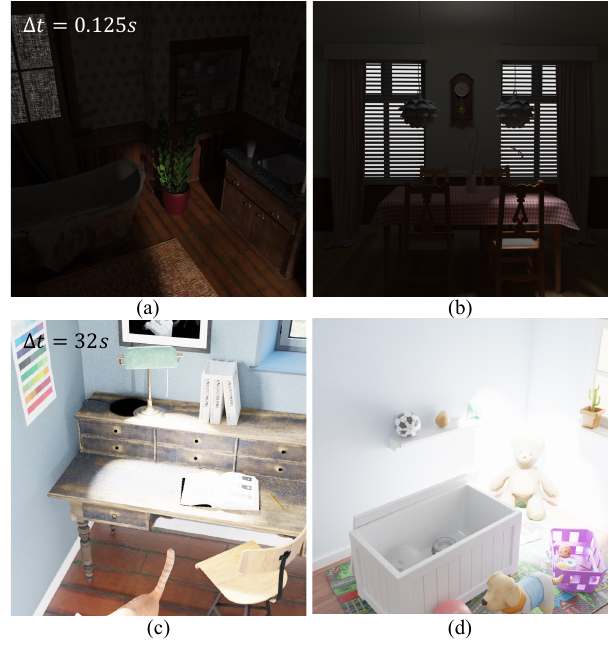}
\vspace{-0.3in}
\caption{
Examples of (a, b) underexposure and (c, d) overexposure. $\Delta t$: Exposure time.
}
\label{dataset_visualization}
\vspace{-0.3cm}
\end{figure}

Compared to common Low Dynamic Range (LDR) imaging, High Dynamic Range (HDR) imaging enables the capture and representation of a broader range of luminance / brightness levels, thereby providing more realistic and visually appealing representations of real-world scenes \cite{cai2024hdr}. It can encompass both the darkest shadows and the brightest highlights within a single frame.
This capability is crucial in a number of fields such as creative media production, photography, virtual reality, and augmented reality that require precise color reproduction, detailed shadow and highlight information, and enhanced visual realism \cite{wang2021deep}.
This enhanced dynamic range not only facilitates more realistic and visually appealing representations of complex scenes 
 \cite{liu2025joint}, but also improves the performance of various computer vision tasks, such as object recognition, scene segmentation, and depth estimation, by providing richer and more detailed visual information \cite{Yan_2023_CVPR}. \par

Novel View Synthesis (NVS) refers to the process of generating new views of a scene from arbitrary viewpoints, given a set of input images captured from different perspectives \cite{duan20244d}. 
This involves understanding and modeling the underlying 3D structure of a scene, as well as accurately rendering the appearance of the scene and objects. Most NVS works focus on LDR image models which fall short in those domains requiring HDR rendering.

Indeed, a couple of recent works \cite{cai2024hdr, huang2022hdr} have studied the HDR-NVS problem by capturing multiple exposures of LDR images per view about the same scene.  However, multiple exposure-based approaches remain vulnerable to motion artifacts, ghosting effects, and demand precise alignment of images captured under varying exposure settings \cite{hdr_reconstruction}. Specifically, longer exposure frames tend to accumulate object or camera movement, leading to blurred details—an issue that becomes more pronounced when exposure durations differ significantly \cite{kalantari2017deep}. During HDR synthesis, pixel-wise fusion (\textit{e.g.}, weighted averaging) can superimpose differing object positions onto the same region, producing semi-transparent or duplicated contours that are especially evident in dynamic scenes with significant object displacement \cite{reinhard2020high}. Furthermore, variations in exposure times often yield discrepancies in brightness distribution, local contrast, and overall appearance, complicating conventional registration algorithms. Finally, in rapidly changing environments or when using mobile devices, capturing multiple exposures in quick succession may prove impractical, limiting the applicability of these methods. \par

To address these issues, we propose a more deployable yet more challenging task, namely {\em HDR-NVS with single-exposure LDR images}, which eliminates the reliance on multiple exposures and thus avoids the aforementioned limitations. However, as illustrated in Fig. \ref{dataset_visualization}, single-exposure images frequently suffer from overexposure or underexposure, posing significant challenges for HDR-NVS.
Furthermore, we present a novel HDR 3D scene modeling framework, {\bf Mono-HDR-3D},
characterized by learning to approximate the underlying LDR image formation process of camera imaging.
Specifically, we start with learning an LDR 3D scene model from single-exposure LDR training images, followed by lifting the color space to HDR with a dedicated color transformation module in a per-channel manner. This from-LDR-to-HDR design is opposite to previous models since single-exposure LDR images are insufficient for deriving an HDR model. We further introduce a closed-loop design by augmenting a process of converting HDR images to LDR images, allowing additional supervision even in the case of no access to HDR ground-truth training data.

Our {\bf contributions} can be summarized as follows.
{\bf(I)} 
We introduce a new HDR-NVS problem where  only single-exposure LDR images are needed so that the data acquisition process is made significantly easier and generic, as well as eliminating those intrinsic limitations with multiple exposures.
{\bf(II)} We propose a generic framework, Mono-HDR-3D, that learns to capture the underlying camera imaging process for bridging LDR and HDR space effectively under the challenging single exposure scenario. Designed as a generic approach, our method can be integrated with different 3D scene models such as NeRF \cite{nerf} or 3D Gaussian Splatting (3DGS) \cite{3dgs}.
{\bf(III)} Extensive experiments validate that our method significantly outperforms previous alternatives.

\begin{figure*}[htbp]
\centering
\includegraphics[width=15.9cm]{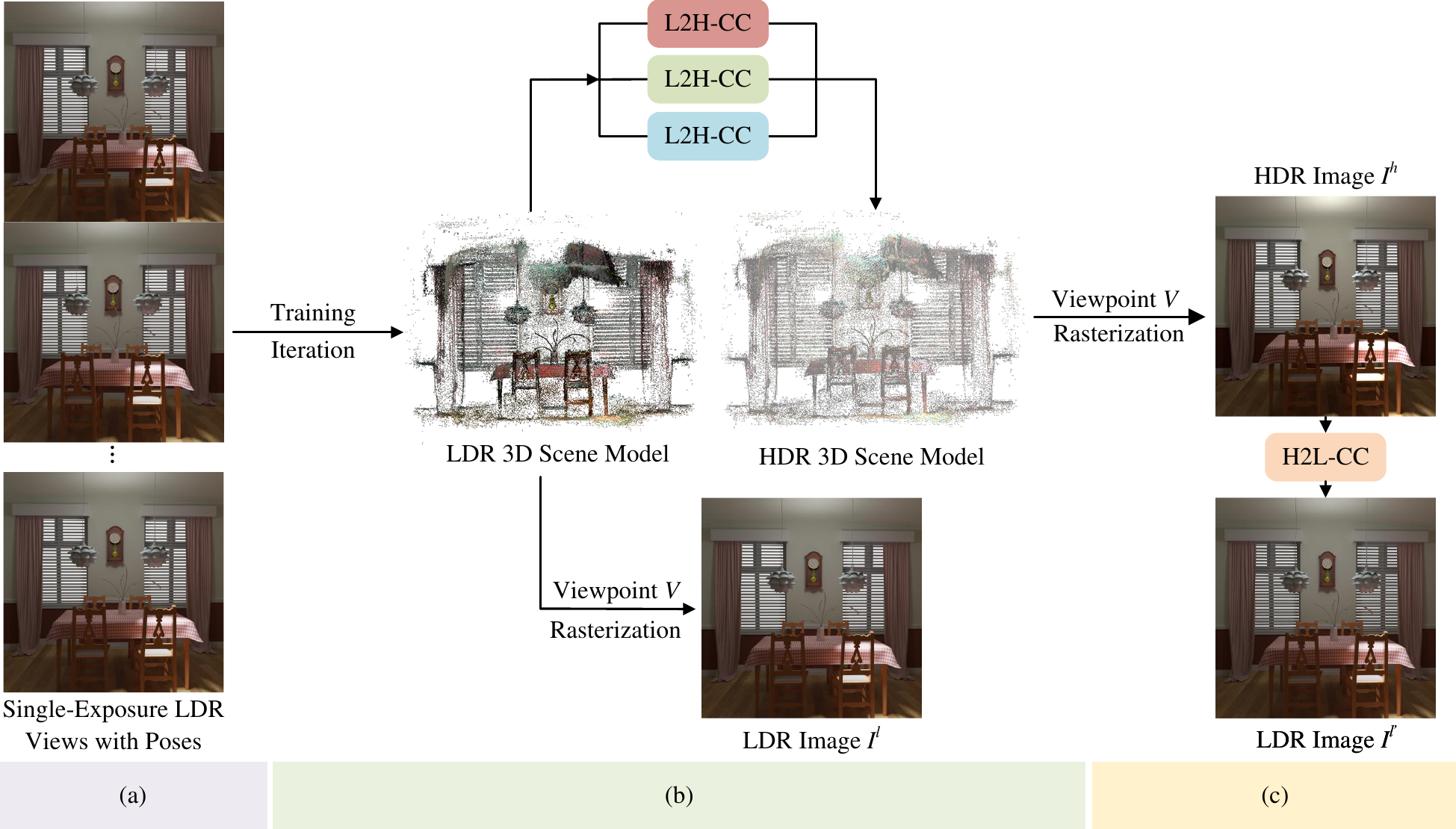}
\caption{Overview of Mono-HDR-3D. (a) Given single exposure LDR training images with camera poses, we learn an LDR 3D scene model (\textit{e.g.}, NeRF or 3DGS).
(b) Importantly, this LDR model is lifted up to an HDR counterpart via a camera imaging aware LDR-to-HDR Color Converter (L2H-CC).
(c) Further, a closed-loop design is formed by converting HDR images back to LDR counterparts with a latent HDR-to-LDR Color Converter (H2L-CC).
This enables optimizing the HDR model even with LDR training images,
particularly useful in case of no access to HDR training data.
During inference, only the HDR or LDR 3D scene model is needed,
taking the novel camera view as the input and outputting the corresponding 
image rendering.
}
\label{pipeline}
\vspace{-0.4cm}
\end{figure*}

\section{Related Work}

\noindent\textbf{High Dynamic Range Imaging} Conventional HDR imaging primarily relies on specialized high-end cameras to capture HDR images \cite{tiwari2015review}. However, the high cost of these cameras renders them inaccessible to general consumers. An alternative approach involves reconstructing HDR images from imagery captured by LDR cameras using algorithms \cite{wang2021deep}. Before NeRF \cite{nerf} was proposed, two primary approaches have been extensively explored. The first generates HDR content by merging multiple LDR images of the same scene taken at varying exposure levels \cite{kalantari2017deep, yan2020deep}. However, the necessity of capturing LDR images with different exposures demands specific software and hardware capabilities, which is not only costly but also brings in various issues, as discussed earlier.
Thus, the second focuses on synthesizing HDR imagery from single-exposure LDR images \cite{eilertsen2017hdr}. Without the challenges associated with multi-exposure capture, it is more feasible for generating HDR images in scenarios where multiple exposures are impractical or datasets are limited \cite{hanji2022comparison}. 

Recent deep learning based methods try to capture the mapping relationship between LDR and HDR images, often achieving state-of-the-art performances \cite{dille2025intrinsic, kim2024dcdr}. 
However, these methods mainly focus on 
individual 2D imagery,
lacking 3D perception capabilities and are not suitable for the novel view HDR image rendering problem. \par

\noindent\textbf{Novel View Synthesis (NVS)} 
is essential for applications such as virtual/augmented reality, gaming, and 3D reconstruction \cite{609457, gao2023nerfneuralradiancefield}. Traditional methods, including Structure-from-Motion (SfM) \cite{schonberger2016structure} and Multi-View Stereo (MVS) \cite{rosu2022neuralmvs}, rely on multi-view geometry to reconstruct 3D scenes but often struggle with occlusions, textureless regions, and high computational costs \cite{jiang2023view}.
Recent advances in NVS leverage deep learning to learn continuous scene representations. Notably, NeRF \cite{nerf} encodes color and density in a neural network, enabling novel view synthesis by querying 3D coordinates and viewing directions. Extensions such as Mip-NeRF \cite{barron2021mip}, FastNeRF \cite{garbin2021fastnerf}, and transformer-based models \cite{lin2023vision, miyato2023gta} have improved efficiency, scalability, and quality. However, most techniques focus on LDR outputs, limiting their applicability for HDR rendering.
Alternatively, 3D Gaussian Splatting (3DGS) represents scenes using learnable 3D Gaussians optimized with multi-view supervision \cite{3dgs} which bypasses volumetric integration and heavy network optimizations, achieving faster training and inference, facilitating real-time rendering. 

\noindent\textbf{HDR Novel View Synthesis (HDR-NVS)} aims to generate novel view HDR images from LDR observations, crucial for scenes with large brightness variations and rich details. 
Huang \textit{et al.}  proposed the first HRD-NVS model, HDR-NeRF \cite{huang2022hdr}, by extending the standard NeRF \cite{nerf} to learn an implicit mapping from physical radiance to HDR color.
However, this method is costly in both model training and inference.
Taking the advantage of 3DGS \cite{3dgs}, \citet{cai2024hdr} addressed this issue
by learning an MLP-based tone-mapper between LDR and HDR models.
Despite their promising results, these methods rely on multiple-exposure LDR training imagery, limiting their applicability in cases with dynamic environments or limited image capturing conditions.
To address this limitation, we introduce single-exposure HDR-NVS, which leverages only single-exposure LDR images. 

\vspace{-0.35cm}
\section{Method}
\label{sec:method}
\vspace{-0.15cm}
\subsection{Problem formulation}
\label{formulation}
\vspace{-0.1cm}
For each of $N$ distinct viewpoints $\boldsymbol{V}=\{V_1, V_2, \cdots, V_N\}$, we capture a set of single-exposure LDR images denoted as $\boldsymbol{I}_{\boldsymbol{V}}^l=\{\boldsymbol{I}^l_{V_1}, \boldsymbol{I}^l_{V_2}, \cdots, \boldsymbol{I}^l_{V_N}\}$.
The objective is to learn a 3D scene model $\mathcal{F}$ that can synthesize an HDR image $\boldsymbol{I}^{h}_{V_{new}}$ for any given novel viewpoint $V_{new}$:
\begin{equation}
    \mathcal{F}: (\boldsymbol{I}_{\boldsymbol{V}}^l, \boldsymbol{V}_{new}) \rightarrow \boldsymbol{I}^{h}_{V_{new}}.
\end{equation}
The synthesized HDR image $\boldsymbol{I}^{h}_{V_{new}}$ needs to exhibit an expanded dynamic range compared to LDR training imagery, while maintaining geometric coherence with the underlying 3D structure of the scene \cite{reinhard2020high}. 
Let $G$ represent the 3D geometry inferred from $\boldsymbol{I}^l$, then $\mathcal{F}(\boldsymbol{I}^l_{\boldsymbol{V}},V_{new})$ must align with $G$ at a viewpoint $V_{new}$. In addition, the HDR synthesis must preserve consistent lighting and color across different views \cite{recover_hdr}. 

Formally, we need to ensure the following constraint holds for each 3D scene point:
\begin{equation}
\resizebox{0.999\hsize}{!}{$C(\boldsymbol{I}^{h}_{V_{new}}(\pi_{V_{new}}(X))) \approx C(\boldsymbol{I}^{l}_{V_{i}}(\pi_{V_{i}}(X))), \forall X\in G, \forall V_i\in \boldsymbol{V}$}
\end{equation} where $C$ denotes the color information of images, $X$ represents the  3D point's coordinates, and $\pi_V(X)$ is a projection function mapping the 3D scene point $X$ onto the 2D image coordinates corresponding to viewpoint $V$. \par
These constraints require the model $\mathcal{F}$ to effectively utilize the limited information from single-exposure inputs to compensate for the absence of multi-exposure sequences. 

\subsection {Mono-HDR-3D}
\label{sec:mono-hdr-3d}
\paragraph{Architecture.} 
To address the proposed problem  as in Sec. \ref{formulation}, we propose a novel single-exposure HDR-NVS framework, \textit{Mono-HDR-3D}. Specifically, given single-exposure LDR images (with corresponding camera poses) as input, we first learn an LDR 3D scene model (\textit{e.g.}, NeRF \cite{nerf} or 3DGS \cite{3dgs}).
This is because single-exposure LDR images provide insufficient information to fully recover an HDR scene.
Then, we elevate this LDR model to an HDR counterpart via our camera-imaging–aware LDR-to-HDR Color Converter (L2H-CC). Additionally, we introduce a latent HDR-to-LDR Color Converter (H2L-CC) as a closed-loop component, enabling the optimization of HDR features even when only LDR training images are available, which ensures the framework to be robust in the absence of ground-truth HDR data. The overall architecture of Mono-HDR-3D is depicted in Fig. \ref{pipeline}.

\paragraph{Camera imaging mechanism.}
We embark with the seminal LDR image formation formula \cite{hasinoff2010noise}:
\begin{equation} 
\label{ldr_formulation}
    I_l = \begin{cases}
        \Delta t/g\cdot I_h + I_0 + \epsilon, & \text{Unsaturation;} \\
        I_{\text{max},} & \text{Saturation}
    \end{cases}
\end{equation}
where $I_l$ denotes the LDR color, $\Delta t$ is the exposure time, $g$ is the sensor gain, $I_h$ represents the corresponding HDR pixel value, and $I_0$ is the constant offset current with $\epsilon$ denoting the sensor noise. Unsaturation refers to those pixels that can be accurately represented by the LDR image after the camera's imaging pipeline processing, while saturation occurs when the sensor reaches its limit, causing the pixel value to be capped at a maximum saturation value $I_{\text{max}}$.

Let the saturated pixel value in Eq. (\ref{ldr_formulation}) of LDR images as
\begin{equation}
I_{\text{max}} = I_{\text{ideal}} - I_{\text{overflow}}
\label{eq:aux}
\end{equation}
where $I_{\text{ideal}}$ and $I_{\text{overflow}}$ represent the pixel values captured by an infinitely capable camera and the overflow values between the ideal and real cameras, respectively. For unsaturation pixels, obviously $I_{\text{overflow}}=0$. 

By integrating Eq. (\ref{eq:aux}) with Eq. (\ref{ldr_formulation}), the formation process of LDR images can be unified as:

\begin{equation}
\begin{aligned}
    I_l & = \underbrace{\Delta t/g\cdot I_h}_{}\underbrace{+ I_0 + \epsilon - I_{\text{overflow}}}_{}, \\
    & \hspace{0.9cm} D(\cdot) \hspace{1.4cm} B(\cdot)
\end{aligned}
\label{eq:hdr2ldr}
\end{equation}
where the term $D(\cdot)$ is responsible for linearly scaling the brightness values of HDR images to fit within the representation range of LDR images, while the term $B(\cdot)$ is to learn the offset and correction of LDR image brightness values.

By reversing Eq. (\ref{eq:hdr2ldr}), the HDR value can be obtained as:

\begin{equation}
\begin{aligned}
    I_h & = \underbrace{g/\Delta t}_{}\cdot\underbrace{(I_l - I_0 + I_{\text{overflow}})}_{} \underbrace{-g/\Delta t\cdot\epsilon}_{}, \\
    & \hspace{0.5cm} X(\cdot) \hspace{1.4cm} S(\cdot) \hspace{1.6cm}  Y(\cdot)
\end{aligned}
\label{eq:ldr2hdr}
\end{equation}
where the term $X(\cdot)$ serves as a scaling factor that linearly amplifies the brightness values of the LDR image to match the range of HDR images,
the term $S(\cdot)$ adjusts and corrects the amplified LDR brightness values,
and the term $Y(\cdot)$ performs noise correction on the adjusted HDR brightness values.

\begin{figure}[t]
\centering
\includegraphics[scale=0.58]{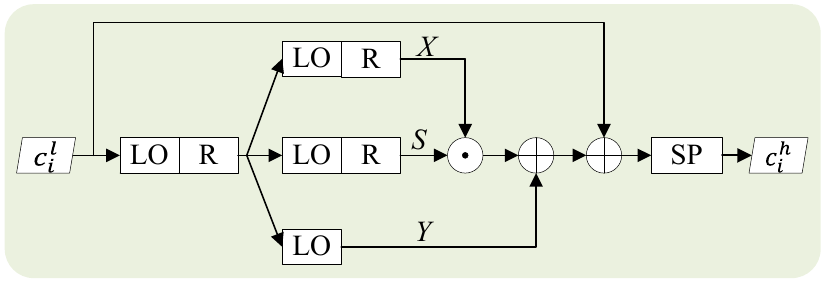}
\caption{Structure of our camera imaging aware LDR-to-HDR Color Converter (L2H-CC).
$c_i^l$/$c_i^h$: LDR/HDR color;
LO: Linear Operation,
R: ReLU,
SP: Softplus.
$\odot$ and $\oplus$: Element-wise multiplication and addition.
}
\label{pl:l2h-cc}
\end{figure}

\paragraph{\MF.}
Simulating the above camera imaging formula Eq. (\ref{eq:ldr2hdr}), we design an LDR-to-HDR color converter, \MF, that learns to approximate the inherent camera response characteristics and facilitates accurate HDR color estimation:
\begin{equation}
    \boldsymbol{c}_i^h = \boldsymbol{f_{\text{l2h}}}(\boldsymbol{c}_i^l),
\end{equation}
where $\boldsymbol{c}_i^h$ and $\boldsymbol{c}_i^l$ represent the HDR color and the LDR color, respectively. 
This is challenging as only the LDR color $I_l$ is known whilst all the rest are not,
resulting in vast modeling freedom.

To address this challenge, we impose network architectural prior in the spirit of camera imaging. That being said, \MF~ consists of three dedicated modules organized in a way that approximates the camera's color conversion behavior (Eq. \eqref{eq:ldr2hdr}), as shown in Fig. \ref{pl:l2h-cc}.
Given an LDR color $c^l_i$, a linear layer with a ReLU activation is first used to convert LDR colors into a latent feature space.
To simulate the three terms $S(\cdot)$, $X(\cdot)$ and $Y(\cdot)$, 
we adopt a simple MLP with ReLU for efficient non-linear computation.
The ReLU activation ensures nonnegative outputs, aligning with the underlying physical constraints of these parameters \cite{hasinoff2010noise}.
Note that no activation function is applied to the $Y(\cdot)$ module, as the noise component $\epsilon$ is inherently random. We also adopt a residual structure \cite{resnet}, which stabilizes the learning process by capturing subtle discrepancies between the LDR input and the HDR output, thereby preserving fine-grained color details more effectively.

\begin{figure}[t]
\centering
\includegraphics[scale=0.87]{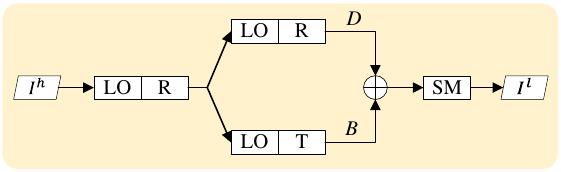}
\vspace{-0.4cm}
\caption{
Structure of our camera imaging aware HDR-to-LDR Color Converter (H2L-CC). $\boldsymbol{I}^h$/$\boldsymbol{I}^l$: HDR/LDR image;
LO: Linear Operation,
R: ReLU, T: Tanh,
SM: Sigmoid.
$\oplus$: Element-wise addition.
}
\label{pl:h2l-cc}
\end{figure}

\paragraph{\MS.} 
We further introduce a closed-loop design, \MS, that converts the rendered HDR images back to LDR for enabling HDR model optimization even when only LDR training data is available.
Similarly, we formulate this component according to the camera imaging principle expressed in Eq. (\ref{eq:hdr2ldr}), formally denoted as:
\begin{equation}
    \boldsymbol{I}^l = \boldsymbol{f_{\text{h2l}}}(\boldsymbol{I}^h),
\end{equation}
where $\boldsymbol{I}^l$ and $\boldsymbol{I}^h$ denote the rendered LDR and HDR images, respectively. \par
Concretely, \MS~ is composed of two modules that approximate the terms of Eq. (\ref{eq:hdr2ldr}), as shown in Fig. \ref{pl:h2l-cc}.
We first transform the HDR image colors into a latent feature space with a linear layer followed by ReLU activation. To simulate each term $D(\cdot)$ and $B(\cdot)$, a specific linear layer with activation is employed, with ReLU for $D(\cdot)$ and Tanh for $B(\cdot)$. This choice of activation functions is made by their physical meanings \cite{hasinoff2010noise} as discussed earlier, ensuring that the network can effectively simulate the HDR-to-LDR conversion process.

\subsection{Model optimization and instantiation}
The overall objective loss function of Mono-HDR-3D can be generally expressed as 
\begin{equation}
    \mathcal{L}=\mathcal{L}_{\text{ldr}}+ \alpha\mathcal{L}_{\text{hdr}} + \beta\mathcal{L}_{\text{h2l}},
\label{loss:overall}
\end{equation} 
where $\mathcal{L}_{\text{ldr}}$ denotes the standard loss function 
of the underlying 3D representation model used (\textit{e.g.}, NeRF \cite{nerf} or 3DGS \cite{3dgs}), 
$\mathcal{L}_{\text{hdr}}$ for matching the HDR ground-truth images if available, and $\mathcal{L}_{\text{h2l}}$ is used to train the proposed \MS{} in the same function as $\mathcal{L}_{\text{ldr}}$.
The two hyper-parameters, $\alpha$ and $\beta$, control the relative importance among the terms.

{\bf Mono-HDR-GS} is obtained by integrating Mono-HDR-3D with 3DGS \cite{3dgs}. For the LDR branch, we adopt:
\begin{equation}
    \mathcal{L}_{\text{ldr}}=\mathcal{L}_1(\boldsymbol{I}^l, \boldsymbol{\hat{I}}^l)+\lambda \cdot \mathcal{L}_{\text{D-SSIM}}(\boldsymbol{I}^l, \boldsymbol{\hat{I}}^l),
\label{loss:ldr}
\end{equation} 
where the $\mathcal{L}_1$ loss and D-SSIM loss \cite{2004Image} are balanced by $\lambda$. $\boldsymbol{\hat{I}}^l$ denotes the ground-truth LDR images.
To optimize HDR generation, following HDR-GS we use a $\mathcal{L}_2$ loss in the $\mu$-law LDR~\cite{kalantari2017deep} domain as 
\begin{equation}
\resizebox{0.99\hsize}{!}{$\mathcal{L}_{\text{hdr}}=\parallel 
    \frac{\text{log}(1+\mu\cdot \text{norm}(\boldsymbol{I}^h))}{\text{log}(1+\mu)} - \frac{\text{log}(1+\mu\cdot \text{norm}(\boldsymbol{\hat{I}}^h))}{\text{log}(1+\mu)}
    \parallel_2^2$},
\label{loss:hdr}
\end{equation} 
where $\mu$ denotes the amount of compression, $\boldsymbol{\hat{I}}^h$ represents the ground-truth HDR images, and \text{norm}$(\cdot)$ specifies the min-max normalization. \par

{\bf Mono-HDR-NeRF} is formed by integrating Mono-HDR-3D with NeRF \cite{nerf}. 
In this case, we adopt the Mean Square Error (MSE) based loss function same as HDR-NeRF:
\begin{equation}
    \mathcal{L}_{\text{ldr}}=\mathcal{L}_{\text{hdr}}=\mathcal{L}_{\text{h2l}}=
    \mathbf{MSE}(\boldsymbol{I}^l, \boldsymbol{\hat{I}}^l).
\end{equation}

\begin{table*}[ht]
\caption{
Quantitative results on the synthetic datasets. For the LDR results, we report averaged across exposure times $t_{1}$, $t_{3}$, and $t_{5}$. All results are averaged over all scenes.
}
\vspace{0.1in}
\centering
\renewcommand{\arraystretch}{1.0}
{%
    \begin{tabular}{lccccccc}
        \toprule[0.15em]
        \multirow{2}{*}{Method} 
        & \multirow{2}{*}{Inference Speed (fps)} 
        & \multicolumn{3}{c}{LDR result} 
        & \multicolumn{3}{c}{HDR result} \\
        \cmidrule(lr){3-5} \cmidrule(lr){6-8}
          & 
          & PSNR$\uparrow$ & SSIM$\uparrow$ & LPIPS$\downarrow$ 
          & PSNR$\uparrow$ & SSIM$\uparrow$ & LPIPS$\downarrow$ \\
        \midrule[0.1em]
        HDR-NeRF          &\bf 0.26 & 30.62 & 0.658 & 0.285 & 13.76 & 0.511 & 0.443 \\ 
        Mono-HDR-NeRF~(Ours)         &\bf 0.26 & \textbf{38.78} & \textbf{0.936} & \textbf{0.048} & \textbf{32.86} & \textbf{0.940} & \textbf{0.068} \\
        \midrule[0.1em]
        HDR-GS         &\bf 147.45 & 39.48 & 0.977 & 0.018 & 35.30 & 0.965 & 0.030 \\
        Mono-HDR-GS (Ours)           & 136.97 & \textbf{41.68} & \textbf{0.983} & \textbf{0.009} &\bf 38.57 & \textbf{0.975} & \textbf{0.012} \\
        \bottomrule[0.15em]
    \end{tabular}%
}
\label{tb:syn}
\end{table*}

\begin{figure*}[htbp]
\centering
\includegraphics[width=16.8cm]{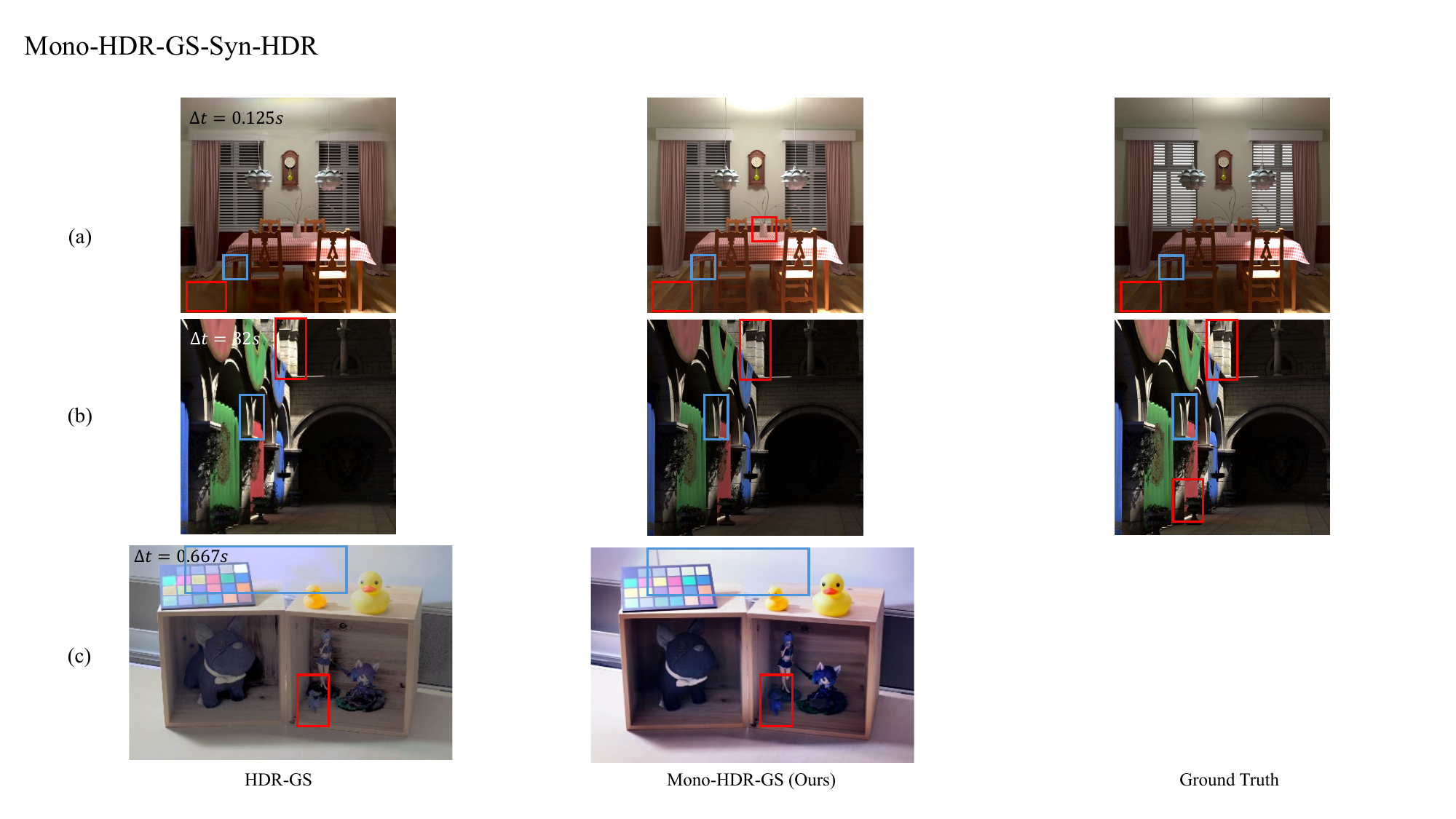}
\caption{Comparison of HDR NVS on both ({a/b}) synthetic and ({c}) real datasets.   
$\Delta t$: Exposure time.
}
\label{fig:gs_syn_hdr}
\end{figure*}
\section{Experiments}
\textbf{Datasets.} Following HDR-GS and HDR-NeRF, we use the multi-view image dataset with 8 synthetic scenes created by the software Blender \cite{blender} and 4 real scenes captured by a camera, where each scene contains 35 images captured under 5 different exposure times $\{t_1, t_2, t_3, t_4, t_5\}$. We use the same training and test data, where images at 18 views with the exposure time randomly selected from $\{t_1, t_3, t_5\}$ are used for training, while the other 17 views at the same exposure time and HDR images are used for testing. Under our proposed single exposure setting, {\em only a specific exposure time} is selected for model training and evaluation for one experiment.
All methods are compared fairly using the same training and test sets.

\textbf{Evaluation metrics.}
We employ the PSNR and SSIM as quantitative metrics.
We utilize LPIPS as a perceptual metric, where lower values signify better perceptual quality. 
Similar to HDR-GS \cite{cai2024hdr}, we also quantitatively evaluate the rendered HDR images in the tone-mapped domain and qualitatively show HDR results tone-mapped by Photomatrix Pro \cite{photomatrix_pro}.

\textbf{Implementation details.} 
Both models are trained with the Adam optimizer with the same parameters as HDR-NeRF and HDR-GS.
For Eq. \eqref{loss:overall}, we set $\beta$ to 0.01/0.05 , while $\alpha=0.6$ for Mono-HDR-NeRF/Mono-HDR-GS.
We set the learning rate of \MF/\MS{} to $5\times 10^{-4}$/$1\times 10^{-3}$,
and the decays to $5\times 10^{-5}$/$5\times 10^{-4}$ by cross-validation.
\par

\subsection{Quantitative evaluation}
\textbf{Competitors.} We compare Mono-HDR-3D with two latest state-of-the-art approaches: (1) HDR-NeRF \cite{huang2022hdr}, the first to synthesize HDR images of novel views using the implicit NeRF model,
and 
(2) 
HDR-GS \cite{cai2024hdr}, which leverages the efficient representation of 3DGS to build an HDR representation model.
To the best of our knowledge, these are the only existing methods specifically designed to synthesize HDR novel views from LDR training imagery.
Whilst designed for multi-exposure LDR setting, they can be also applied
to our proposed single-exposure setting.
We used their official repositories to conduct the experiments
for ensuring their own optimal performance.

Tab. \ref{tb:syn} presents the quantitative results on the synthetic datasets for both LDR and HDR NVS. Notably, HDR NVS results are the most important, as they encapsulate the core objective of this study. The reported results are averaged across three different exposure times to ensure the completeness and reliability of the performance metrics.

\begin{figure*}[htbp]
\centering
\includegraphics[width=16.6cm]{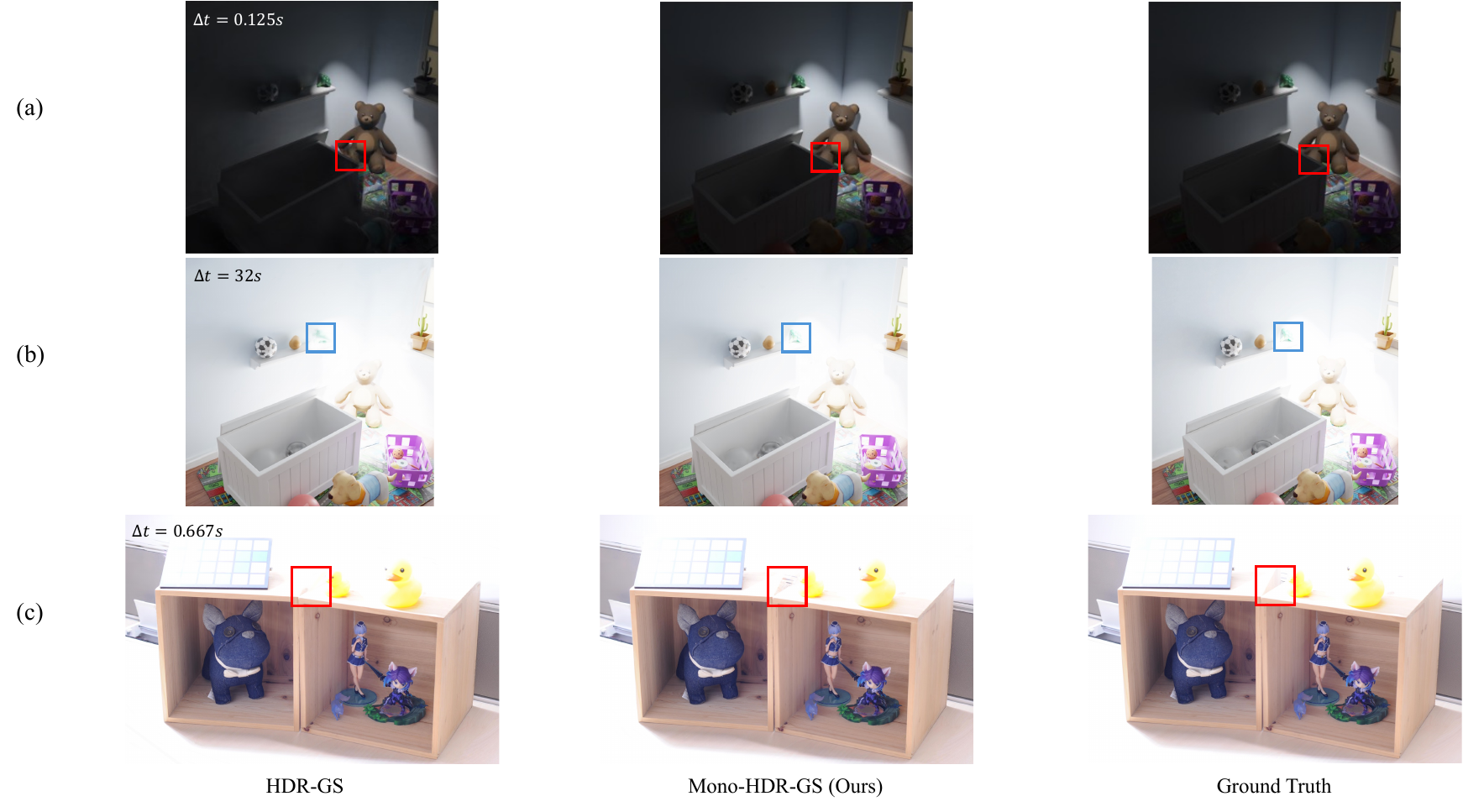}
\vspace{-0.3cm}
\caption{Comparison of LDR NVS on both ({a/b}) synthetic and ({c}) real datasets. 
$\Delta t$: Exposure time.}
\label{fig:gs_syn_ldr}
\end{figure*}

In addition to visual quality assessment, model inference speed (fps) is also included. We highlight the following key points:

(I) In terms of HDR NVS results, our models significantly outperform all alternatives, particularly HDR-NeRF, in generation quality. This advantage arises because HDR-NeRF struggles to converge without multiple exposure LDR training data, often producing images that are entirely black or white. This highlights the greater challenges associated with the proposed single exposure setting while also validating the efficacy and superiority of our model design in addressing such challenges.

(II)
Regarding the LDR NVS results, we observe a similar performance advantage with our models. This indicates that directly learning an HDR model from single-exposure LDR data, as competitors do, would be inferior due to the absence of multiple exposure observations. This also partly explains our stronger HDR NVS results, which somehow are dependent on the quality of the LDR output. 

(III)
In terms of efficiency, our models perform comparably to alternatives using either NeRF or 3DGS as the representation model. This suggests that our models do not sacrifice efficiency for the sake of quality.

\textbf{Result analysis on real data.}
While less important, Tab. \ref{tb:real} presents the quantitative results of LDR NVS on the real datasets, as there are no ground-truth HDR images available. The results are averaged across three distinct exposure times and encompass all scenes. It is evident that no method clearly stands out in synthesis quality.

\begin{table}[ht]
\caption{Quantitative results on the real datasets. We report the results averaged across all scenes and exposure times $t_{1}$, $t_{3}$, and $t_{5}$.
Note, {\em HDR results cannot be reported due to no ground-truth.}
}
\vspace{0.1in}
\centering
\renewcommand{\arraystretch}{1.0}
\resizebox{\columnwidth}{!}
{
    \begin{tabular}{lccc}
    \toprule[0.15em]
    \multirow{2}{*}{Method} & \multicolumn{3}{c}{LDR result} \\
    &PSNR$\uparrow$ & SSIM$\uparrow$ & LPIPS$\downarrow$ \\
    \midrule[0.1em]
    HDR-NeRF                     & 32.50       &\bf 0.948    &\bf 0.069  \\
    Mono-HDR-NeRF~(Ours)         &\bf 32.52    &\bf 0.948       &\bf 0.069 \\
    \midrule[0.1em]
    HDR-GS                       & 35.34    & 0.966    & 0.019  \\
    Mono-HDR-GS (Ours)           &\bf 35.81 &\bf 0.967 &\bf 0.017  \\
    \bottomrule[0.15em]
\end{tabular}}
\vspace{-0.6cm}
\label{tb:real}
\end{table}

\subsection{Qualitative evaluation}
\label{sec:exps:qualitative_evaluation}
Numerical metrics such as PSNR, SSIM, and LPIPS may not fully reflect the perceived quality of images. Therefore, a qualitative evaluation through visual comparison is essential.
For HDR NVS results on both synthetic and real datasets, as shown in Fig. \ref{fig:gs_syn_hdr}, HDR-GS struggles to accurately reconstruct the darkest and brightest details, whereas our Mono-HDR-GS excels in rendering more intricate structures. Regarding LDR NVS results, as illustrated in Fig. \ref{fig:gs_syn_ldr}, HDR-GS tends to produce blurry and visually unappealing results (\textit{e.g.}, the synthetic data case) or fails in rendering extremely bright and contrastive regions (\textit{e.g.}, the real data case). In contrast, our Mono-HDR-GS can successfully recover smoother color details and present the brightness properly for such challenging cases. \par 
\begin{figure*}[htbp]
\centering
\includegraphics[height=5.5cm]
{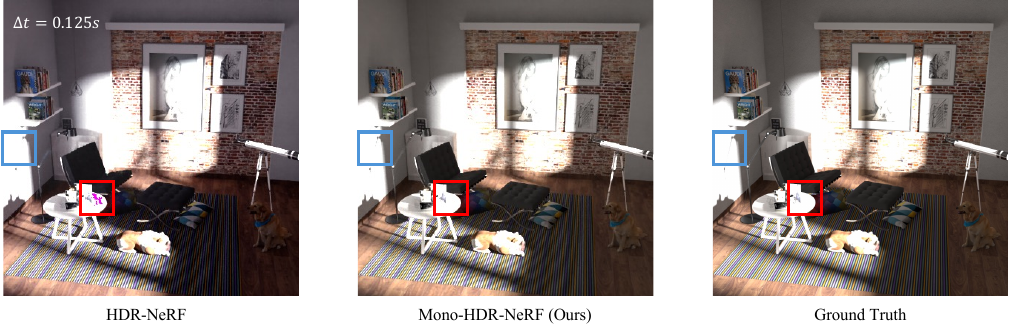} 
\vspace{-0.1in}
\caption{HDR reconstruction comparison on synthetic datasets. $\Delta t$: Exposure time.}
\label{fig:nerf_syn_hdr}
\end{figure*}

We make similar observations when comparing NeRF-based models, as reflected in Fig. \ref{fig:nerf_syn_hdr} (Also see Fig. \ref{fig:nerf_all_ldr} in Appendix \ref{app:nerf_vis}), where HDR-NeRF produces color artifacts and blurry outputs, while Mono-HDR-NeRF achieves superior color consistency and detail preservation.


\begin{table}[tbp]
\caption{
Ablation analysis on the synthetic datasets.
The results are averaged across exposures and scenes.
\texttt{MLP}: Alternative design with similar amount of parameters.
}
\vspace{0.1in}
\centering
\renewcommand{\arraystretch}{1.0}
\resizebox{0.45\textwidth}{!}{%
    \begin{tabular}{cccccccc}
        \toprule[0.15em]
        \multirow{2}{*}{Row} 
        & \multicolumn{2}{c}{\text{\MF}} 
        & \multicolumn{2}{c}{\text{\MS}} 
        & \multicolumn{3}{c}{\text{HDR result}} \\
        \cmidrule(lr){2-3} \cmidrule(lr){4-5} \cmidrule(lr){6-8}
        & MLP & Ours & MLP & Ours & PSNR$\uparrow$ & SSIM$\uparrow$ & LPIPS$\downarrow$ \\
        \midrule[0.1em]
        1 & $\surd$  &          &         & $\surd$  & 19.02 & 0.778 & 0.327 \\
        2 &          & $\surd$  & $\surd$ &          & 38.43 & 0.974 & 0.015 \\
        3 &          & $\surd$  &         & $\surd$  &\bf 38.57 &\bf 0.975 &\bf 0.012 \\
        \bottomrule[0.15em]
    \end{tabular}%
}
\vspace{-0.5cm}
\label{tb:ablation:module_design}
\end{table}

\begin{table}[ht]
\caption{
Ablation analysis of closed-loop design on the synthetic datasets.
The results are averaged across exposures and scenes.
}
\vspace{0.1in}
\centering
\renewcommand{\arraystretch}{0.5}
\resizebox{0.45\textwidth}{!}
{%
    \begin{tabular}{ccccccc}
        \toprule[0.15em]
        \multirow{2}{*}{Row} 
        & \multirow{2}{*}{\MF} 
        & \multirow{2}{*}{\MS} 
        & \multicolumn{3}{c}{HDR result} \\
        \cmidrule(lr){4-6} 
         & & & PSNR$\uparrow$ & SSIM$\uparrow$ & LPIPS$\downarrow$ \\
        \midrule[0.1em]
        1 & $\surd$    & $\times$ & 38.19 & 0.974 & 0.015  \\
        2 & $\surd$    & $\surd$  &\bf 38.57 &\bf 0.975 &\bf 0.012  \\
        \bottomrule[0.15em]
    \end{tabular}%
}
\label{tb:ablation:closed_loop}
\vspace{-0.3cm}
\end{table}

\subsection{Ablation studies}
We conduct an ablation study with the most efficient model, Mono-HDR-GS, on the synthetic datasets.

\paragraph{Module design.}
To evaluate the design of the proposed \MF~ and \MS~ modules, we compare them with an alternative MLP with a similar number of parameters. 
We make several observations from Tab. \ref{tb:ablation:module_design}:
(I) \textbf{Row 1 vs. 3}: With a plain MLP to replace L2H-CC, the 
model performance will degrade significantly, validating the importance of our simulating the camera imaging process (see Sec. \ref{sec:mono-hdr-3d}).
(II) \textbf{Row 2 vs. 3}: When replacing H2L-CC with MLP, we also observe a performance drop, although quite slight, suggesting that our closed-loop design is useful even when HDR ground truth is available.

\paragraph{Effect of closed-loop design.}
Following the above design based ablation, we further look into the effect of our closed-loop design with H2L-CC with HDR training data.
The results in Tab. \ref{tb:ablation:closed_loop} indicate that this design brings a positive impact of 0.38 dB increase in PSNR for HDR NVS, demonstrating that the closed-loop framework contributes significantly to enhancing the quality of the reconstructed HDR images.

\paragraph{Loss contributions.} 
Based on our proposed Mono-HDR-GS, we conduct systematic ablation studies to evaluate the contribution of each loss component in Eq. \eqref{loss:overall}. The results in Tab. \ref{tb:ablation:losses} reveal three key observations:
(I) \textbf{Row 2 vs. Row 7:} HDR loss $L_\text{hdr}$ serves as the foundational component for HDR-NVS performance. When used alone, Mono-HDR-GS achieves moderate metrics (33.93 dB PSNR, 0.925 SSIM), while its absence leads to severe degradation (\textit{e.g.}, Row 5 shows 13.50 dB PSNR without $L_\text{hdr}$). This validates its critical role in reconstructing high dynamic range scenes. (II) \textbf{Row 4 vs. Row 6:} 
LDR loss $L_\text{ldr}$ provides essential regularization for scene modeling. When combined with $L_\text{hdr}$, it improves PSNR by +4.26 dB (Row 2 vs. 4) and maintains structural fidelity. Notably, $L_\text{ldr}$ alone fails to train (Row 1), but acts as a complementary constraint when paired with HDR-aware objectives. (III) \textbf{Row 4 vs. 7:} Closed-loop loss $L_\text{h2l}$ enhances photometric consistency across exposure domains. Its integration elevates PSNR by +0.38 dB  while reducing perceptual distance. This demonstrates that bidirectional optimization between HDR and LDR spaces refines both radiance estimation and tone mapping.

\begin{table}[tbp]
\caption{
Ablation analysis of different loss components on the synthetic datasets.
Results are averaged across exposures and scenes.}
\vspace{0.5em}
\centering
\renewcommand{\arraystretch}{1.0}
\resizebox{0.40\textwidth}{!}{%
    \begin{tabular}{ccccc}
        \toprule[0.15em]
        \multirow{2}{*}{Row} 
        	& \multirow{2}{*}{Loss}
        & \multicolumn{3}{c}{\text{HDR result}} \\
        \cmidrule(lr){3-5} 
        &  & PSNR$\uparrow$ & SSIM$\uparrow$ & LPIPS$\downarrow$ \\
        \midrule[0.1em]
        1 &  $L_\text{ldr}$  & - & - & - \\
        2 &  $L_\text{hdr}$  & 33.93 & 0.925 & 0.050 \\
        3 &  $L_\text{h2l}$  & 11.87 & 0.504 & 0.371 \\
        	4 & $L_\text{ldr} + L_\text{hdr}$ & 38.19 & 0.974 & 0.015 \\
        	5 & $L_\text{ldr} + L_\text{h2l}$ & 13.50 & 0.507 & 0.359 \\
        	6 & $L_\text{hdr} + L_\text{h2l}$ & 33.58 & 0.934 & 0.058 \\        	
        	7 & $L_\text{ldr} + L_\text{hdr} + L_\text{h2l}$ &\bf 38.57 &\bf 0.975 &\bf 0.012 \\
        \bottomrule[0.15em]
    \end{tabular}%
}
\vspace{-0.5cm}
\label{tb:ablation:losses}
\end{table}

\paragraph{Performance of different LDR / HDR ratios.}  To evaluate the robustness of our Mono-HDR-GS framework under varying data availability, we conduct experiments with LDR / HDR ratios ranging from 0/1 (pure HDR) to 5/1 (dominant LDR), as shown in Tab. \ref{tb:ablation:ldr/hdr}. The results demonstrate:

(I) \textbf{Data Efficiency:} Our model maintains strong performance even with sparse HDR data. For instance, at 5/1 ratio, Mono-HDR-GS retains 92.6\% of its peak PSNR (35.51 dB vs. 38.57 dB at 1/1), suggesting effective knowledge transfer from LDR supervision. 

(II) \textbf{LDR Supervision:} When HDR data becomes extremely scarce (5/1 vs. 0/1), the PSNR degradation of Mono-HDR-GS (from 35.51 dB to 33.93 dB) is significantly smaller than HDR-GS (from 34.89 dB to 33.46 dB). This confirms that LDR supervision provides a more robust geometric prior for HDR reconstruction.

(III) \textbf{HDR Criticality:} Pure HDR supervision (0/1) outperforms pure LDR supervision (1/0) by +23.4 dB PSNR, validating the irreplaceable role of HDR data in capturing radiance information. This finding emphasizes HDR's fundamental importance for high-quality HDR novel view synthesis.

(IV) \textbf{Overall Superiority:} Across all ratios, Mono-HDR-GS consistently outperforms HDR-GS with statistically significant margins. At equal computational cost, our Mono-HDR-GS achieves up to +3.27 dB PSNR improvement (1/1 ratio) and maintains superior structural fidelity (SSIM 0.975 vs. 0.965). Notably, even with 100\% LDR data (1/0 ratio), our method generates marginally better results than HDR-GS trained solely on HDR images, demonstrating the effectiveness of closed-loop design.


\begin{table}[tbp]
\caption{Ablation studies of different ratio of LDR / HDR images.}
\vspace{0.5em}
\centering
\renewcommand{\arraystretch}{1.0}
\resizebox{0.45\textwidth}{!}{%
\begin{tabular}{ccccc}
\toprule[0.15em]
\multirow{2}{*}{Method} & \multirow{2}{*}{LDR / HDR} & \multicolumn{3}{c}{HDR result} \\ \cmidrule{3-5}
& & PSNR$\uparrow$ & SSIM$\uparrow$ & LPIPS$\downarrow$ \\
\midrule[0.10em]
HDR-GS & \multirow{2}{*}{1/1} & 35.30 & 0.965 & 0.030 \\
Mono-HDR-GS & & \bf 38.57 & \bf 0.975 & \bf 0.012 \\
\midrule
HDR-GS & \multirow{2}{*}{2/1} & 35.26 & 0.963 & 0.033 \\
Mono-HDR-GS & & 37.97 & \bf 0.975 & 0.013 \\
\midrule
HDR-GS & \multirow{2}{*}{3/1} & 35.16 & 0.958 & 0.035 \\
Mono-HDR-GS & & 37.53 &  0.974 & 0.014 \\
\midrule
HDR-GS & \multirow{2}{*}{5/1} & 34.89 & 0.961 & 0.027 \\
Mono-HDR-GS & & 35.51 & 0.963 & 0.023 \\
\midrule
HDR-GS & \multirow{2}{*}{0/1} & 33.46 & 0.936 & 0.075 \\
Mono-HDR-GS & & 33.93 & 0.925 & 0.050 \\
\midrule
HDR-GS & \multirow{2}{*}{1/0} & 10.51 & 0.503 & 0.350 \\
Mono-HDR-GS & & 13.50 & 0.507 & 0.359 \\
\bottomrule[0.15em]
\end{tabular}%
}
\vspace{-0.5em}
\label{tb:ablation:ldr/hdr}
\end{table} 

\section{Conclusion}
This paper pioneers the Single-Exposure HDR-NVS problem by introducing Mono-HDR-3D, a novel meta-algorithm designed to operate effectively with only single-exposure LDR images during training. Unlike conventional HDR-NVS approaches that rely on multiple-exposure imagery, Mono-HDR-3D addresses critical limitations such as motion artifacts, high capture and storage costs, and the need for precise exposure tuning. This not only enhances applicability but also simplifies deployment in dynamic and rapidly changing scenes.
Extensive experimental evaluations demonstrate that Mono-HDR-3D significantly outperforms existing methods in generative quality under such more challenging conditions. Importantly, the seamless integration capability of Mono-HDR-3D with existing 3D representation models highlights its versatility and potential for widespread adoption, making advanced HDR techniques accessible to a broader audience and even future advancement in representation modeling.

This work opens new avenues for efficient and robust HDR scene modeling, especially in contexts where access to expensive, professional cameras for training data collection is limited or not possible. By democratizing the process of HDR imaging, we empower more individuals and organizations even with limited resources to engage with high-quality imaging technologies.
Future work will focus on further optimizing Mono-HDR-3D and exploring its application across more diverse real-world environments, solidifying its role as a go-to solution or baseline in the evolution of HDR imaging and 3D scene synthesis, while continuing to make these advancements accessible to all.

\section*{Acknowledgments}
This work was supported in part by the Key Research and Development Plan of Jiangsu Province (Industry Foresight and Key Core Technology Project) under Grant No. BE2023008-2.

\section*{Impact Statement}

In this work, we address the high demands for high-quality visual generation and immersion as required in many fields such as VR / AR, entertainment, creative media, broadcasting, TV, and gaming.
This has great potential for democratizing both the academic research often featured with limited resources, as well as related industries with diverse backgrounds and contexts. Unlike existing methods requiring costly multi-exposure LDR imagery capture, our approach enables HDR scene reconstruction from single-exposure images with strong accuracy and realism.

While this research problem and our technology are still in the early stage, potential misuse risks (\textit{e.g.}, malicious content generation) might warrant ethical considerations. We advocate for responsible deployment and emphasize that its benefits in advancing safer, high-fidelity digital environments outweigh foreseeable risks per se.

\bibliography{example_paper}
\bibliographystyle{icml2025}

\newpage
\appendix
\onecolumn
\icmltitle{Appendix for ``High Dynamic Range Novel View Synthesis with Single Exposure"}
\section{Additional Visualization Comparisons of HDR-NeRF and Mono-HDR-NeRF}

\label{app:nerf_vis}
\begin{figure*}[htbp]
\centering
\includegraphics[width=17.0cm]
{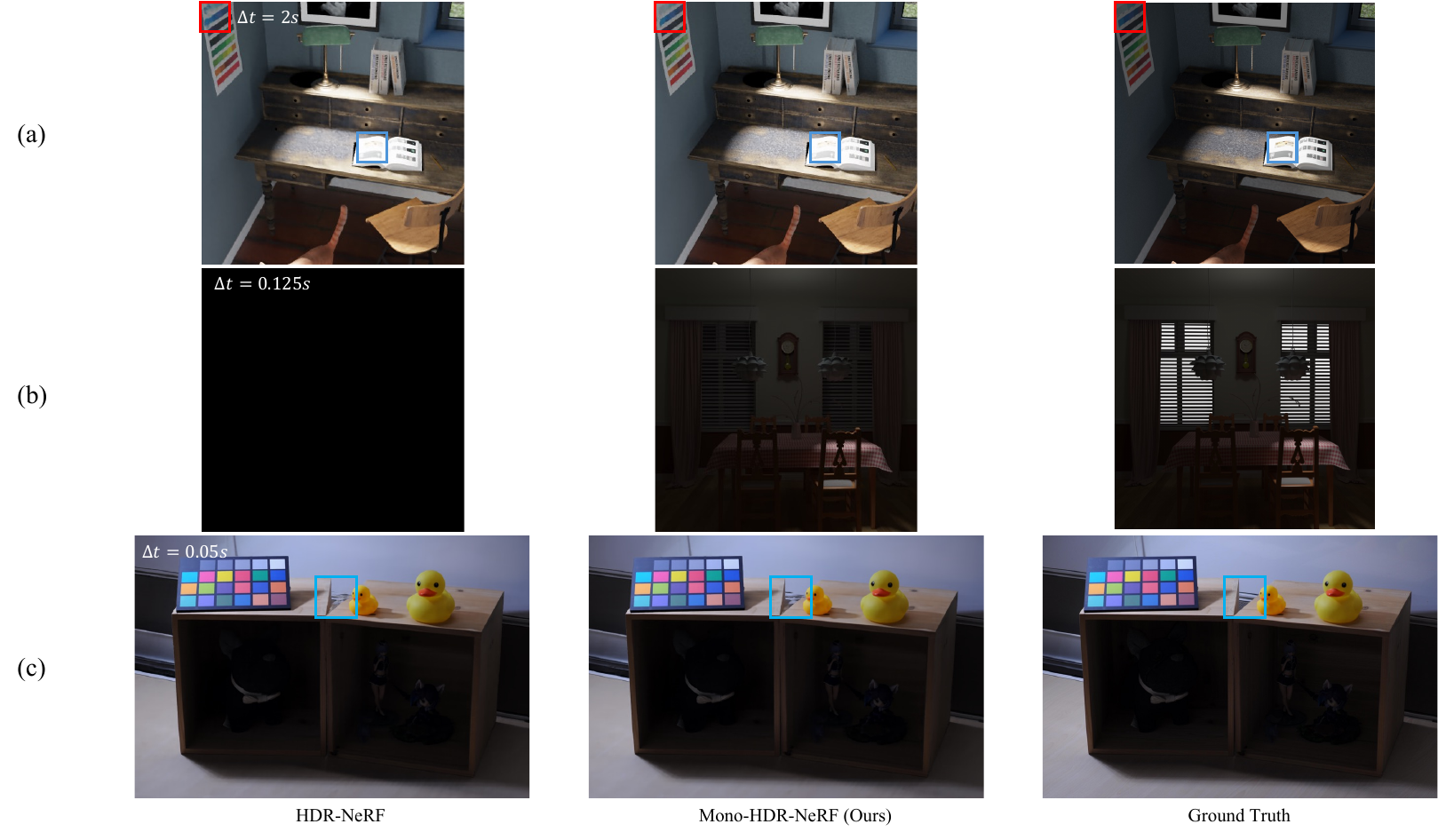} 
\vspace{-0.1in}
\caption{Comparison of LDR NVS on both ({a/b}) synthetic and ({c}) real datasets. 
$\Delta t$: Exposure time.}
\label{fig:nerf_all_ldr}
\end{figure*}

This section presents additional visual comparisons between HDR-NeRF and our Mono-HDR-NeRF, including LDR novel view rendering results on both synthetic and real datasets. These comparisons are illustrated in Fig. \ref{fig:nerf_all_ldr}. 


It can be observed that, HDR-NeRF suffers from color artifacts and blurriness when rendering LDR images, and may fail to converge, producing black outputs without multi-exposure data. In contrast, Mono-HDR-NeRF achieves superior color consistency and detail preservation in LDR rendering.


\end{document}